\documentclass[letterpaper, 10 pt, conference]{ieeeconf}  

\IEEEoverridecommandlockouts                              

\usepackage{amsmath} 
\usepackage{amssymb}  
\usepackage{comment}
\usepackage{graphicx}
\usepackage{xcolor}
\usepackage{tikz}
\usepackage{hyperref}
\usepackage{subcaption}

\usepackage{booktabs, multirow} 

\title{\LARGE \bf A Deep Reinforcement Learning Framework and Methodology for Reducing the Sim-to-Real Gap in ASV Navigation}

\author{Luis F. W. Batista,$^{1,2,*}$ Junghwan Ro,$^{1,*}$ Antoine Richard$^{3}$, Pete Schroepfer$^{1,2,}$, \\ Seth Hutchinson$^{1}$, and Cedric Pradalier$^{2}$
\thanks{This research was supported by the French Agence Nationale de la Recherche, under grant ANR-23-CE23-0030 (project R3AMA).}%
\thanks{$^{1}$ are with Georgia Institute of Technology, Atlanta, USA}%
\thanks{$^{2}$ is with GeorgiaTech Europe - IRL2958 GT-CNRS, Metz, France}%
\thanks{$^{3}$ is with University of Luxembourg, Luxembourg}%
\thanks{$^{*}$ These authors contributed equally to this work.}%
}

\begin{document}

\maketitle
\thispagestyle{empty}
\pagestyle{empty}

\begin{abstract}

Despite the increasing adoption of Deep Reinforcement Learning (DRL) for Autonomous Surface Vehicles (ASVs), there still remain challenges limiting real-world deployment. In this paper, we first integrate buoyancy and hydrodynamics models into a modern Reinforcement Learning framework to reduce training time. Next, we show how system identification coupled with domain randomization improves the RL agent performance and narrows the sim-to-real gap. Real-world experiments for the task of capturing floating waste show that our approach lowers energy consumption by 13.1\% while reducing task completion time by 7.4\%. These findings, supported by sharing our open-source implementation, hold the potential to impact the efficiency and versatility of ASVs, contributing to environmental conservation efforts.

\end{abstract}
\section{Introduction}

Water plastic pollution poses a significant threat to global sustainability, impacting marine life, ecosystems, and human health \cite{van2020plastic}. To restore ecological balance, it is necessary to not only halt pollution but also to reverse its effects. As a solution, Autonomous Surface Vehicles (ASVs) are being increasingly recognized for their critical role in addressing this issue for both cleaning and monitoring water quality \cite{FornaiwaterSampling, chang2021autonomous}. The task of cleaning water surfaces autonomously has gathered attention not only from the academic sphere, \cite{chang2021autonomous, zhou2021time} but also from the industrial sector in companies such as IADYS, Clearbot, WasteShark, and OrcaUboat. While ASVs offer potential solutions for floating waste management, the autonomy of these solutions is often compromised by the complexity of the marine environments. 

The advancement of ASVs depends on the development of robust and efficient control systems capable of handling highly dynamic and often unpredictable marine environments. As these ASVs often need to explore large areas without access to charging points, it is also critical that control algorithms operate efficiently to reduce energy consumption. This reduction in battery energy consumption allows for extended operational range, reduced costs, and lower environmental impact.

At the same time, Deep Reinforcement Learning (DRL) has emerged as a powerful tool in robotics, providing flexible and robust control strategies, and it has growing applications in the field of marine robotics and ASVs \cite{qiao2023survey}. Recent work has even shown that the DRL-based approach is capable of outperforming traditional control algorithms in real-world trials \cite{wang2023deep}. Yet, field applications of DRL for ASVs are scarce, with most studies not extending beyond simulations, especially in the context of energy-efficient agents.

Despite its potential, the adoption of DRL for ASVs is hindered by two significant barriers. Firstly, there is a lack of high-performance marine simulation environments. Current options, such as Gazebo's UUV plugin \cite{bingham2019toward} incorporate foundational buoyancy and hydrodynamics simulations but do not meet the extensive parallelization needs of DRL frameworks. Secondly, there is difficulty in validating and interpreting the internal decision-making processes of RL agents due to their 'black-box' nature. This opaque nature creates uncertainty with respect to RL agent reliability and safety in mission-critical applications, which can hinder the adoption and acceptance of such systems \cite{Lewis2018, XAiRL, SchroepferTrust2023}. 

With these challenges in mind, we aim to narrow the gap between simulation and the real world through methodical system identification coupled with domain randomization. Reducing this gap improves the efficiency of DRL agents for sustainable applications of ASVs, addressing issues such as thruster dynamics, battery charge effects, and external disturbances, thus reducing behavior uncertainty as they are trained under more realistic conditions for real-world deployment. Our work's main contributions include:
\begin{itemize}
    \item Open-source, highly parallelized hydrodynamics and buoyancy implementation suitable for reinforcement learning framework\footnote{\url{https://github.com/luisfelipewb/RL4WasteCapture/}}.
    \item Methodology to reduce the sim-to-real gap by combining System Identification with Domain Randomization for Deep Reinforcement Learning.
    \item Real-world experimental evidence illustrating that our approach can minimize battery energy consumption and simultaneously speed up task completion.
\end{itemize}

\begin{figure}[t]
    \centering
    \begin{subfigure}{0.32\linewidth}
        \includegraphics[width=\linewidth]{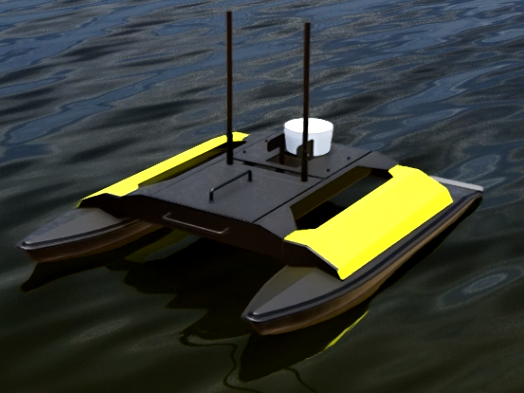}
        \caption{Isaac Sim}
        \label{fig:asv-isaac}
    \end{subfigure}
    \begin{subfigure}{0.32\linewidth}
        \includegraphics[width=\linewidth]{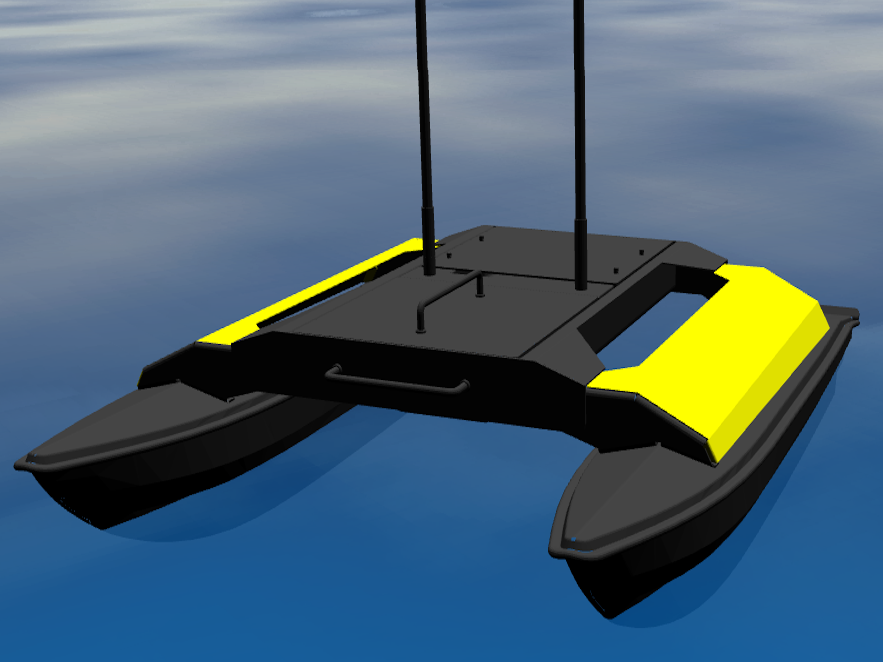}
        \caption{Gazebo}
        \label{fig:asv-gazebo}
    \end{subfigure}
    \begin{subfigure}{0.32\linewidth}
        \includegraphics[width=\linewidth]{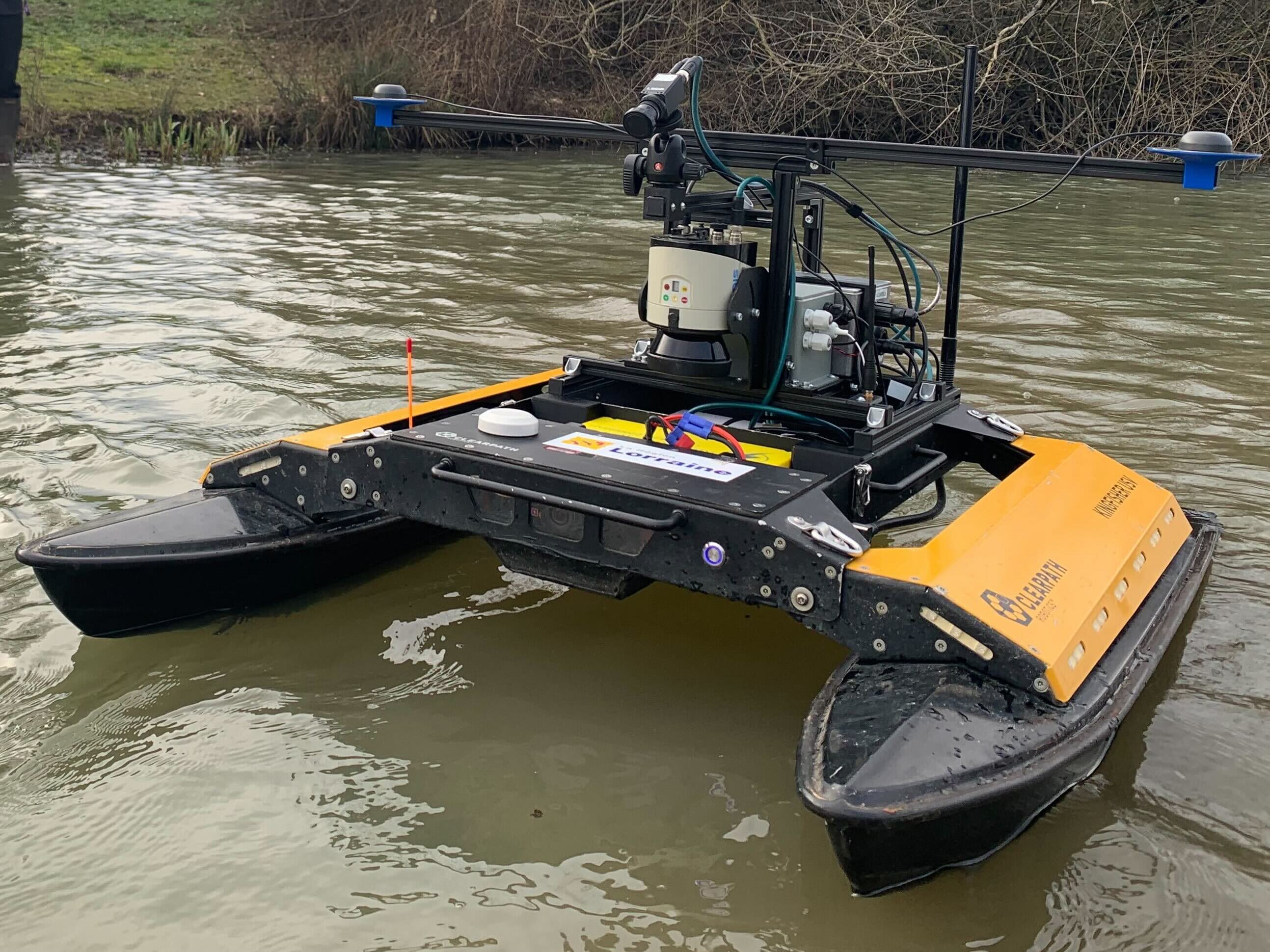}
        \caption{Real ASV}
        \label{fig:asv-real}
    \end{subfigure}
    \caption{Simulated and Real ASV Platforms Used Through our Experiments}
    \label{fig:asv_platform}
\end{figure}

\section{Related Work}
\label{sec:related_work}

There is a growing adoption of Deep Learning (DL) on ASVs in the areas of perception, control, and multi-robot collaboration \cite{qiao2023survey}. Despite this trend, the utilization of DL-based controllers for ASV is still in the early stages. One of the main barriers to adoption lies in the challenge of developing agents capable of handling the complex hydrodynamic water-body interaction \cite{qiao2023survey}.

To address the complex dynamics, DRL-based controllers have been explored as a potential solution. In \cite{zhao2020path}, a DRL-based controller was used for path following. Here, \cite{zhao2020path} demonstrated that in simulation, a DRL-based controller was capable of controlling ASVs in extremely complex systems. Similarly, \cite{lin2023robust} proposes an IQN-based local path planner to compensate for unknown currents and avoid obstacles. RL has also been proposed for ASV control problems in several other studies \cite{zhang2021model, wang2020reinforcement, wang2022reinforcement}. However,  evaluation in these studies is often performed only on simulation and lacks a field test validation, which would demonstrate the ability to move from a training environment to the real world. 

The studies that include the evaluation of DRL agents in real-world scenarios are slowly growing, but more are needed in different subdomains of DRL agent task performance. Wang et al. claim that DRL outperforms NMPC in trajectory tracking, showing lower tracking error and better disturbance rejection in river environments \cite{wang2023deep}. An RL-based strategy to integrate guidance and heading control, improving tracking accuracy and robustness of the controller is proposed in \cite{wang2023path}. While this existing literature shows promising results, most of the research is still in the early stages and often focuses on the path-tracking task. As such, further real-world testing and validation are necessary to continue to fill the research gaps regarding how DRL controllers perform outside of the simulated training environment, in particular, on how changes in the training environment or algorithms can improve real-world operation performances.


While simulation environments exist, the options are rather limited. The most up-to-date option that is under active development is detailed in \cite{bingham2019toward}. They extend previous initiatives to simulate buoyancy and hydrodynamics \cite{manhaes2016uuv} and provide a solid starting point. Cie{\'s}lak introduced Stonefish  \cite{cieslak2019stonefish}, an advanced simulation tool capable of real-time performance. However, it lacks GPU acceleration, limiting its scalability for large-scale simulations required for training DRL agents. Such frameworks suitable for the RL pipeline are available in other domains. For instance, \cite{el2023drift} solves similar navigation tasks for satellites, but it misses the hydrodynamics and buoyancy simulation.


In summary, while the existing literature provides a solid foundation, it could be further enhanced by the inclusion of accessible open-source implementations, comprehensive analysis, and robust field test validation. Our work aims to enrich the field by offering accessible code and evaluating the impact of system identification to minimize the sim-to-real gap, including field test validation.
\section{Problem Formulation}

\subsection{Capture task}

Capturing floating waste requires navigating the ASV over the waste, where a fixed net between its hulls traps it during passage. The target position is provided in a two-dimensional space, and the agent must be capable of navigating toward it by controlling the actions of each thruster. The goal is considered achieved when the ASV approaches the target head-on and passes over it with a tolerance of up to $0.3\ m$ distance between the target's position and the center of mass (COM) of the vessel. We considered an environment free of obstacles to isolate the evaluation of the capture task. Anticipating future integration of perception capabilities of the onboard camera, the task has been designed to reflect a realistic field of view of 90 degrees and capping the operational range at 10 meters.

\subsection{Dynamic Model}

The dynamic model of the ASV in the simulation is based on Fossen's six degrees of freedom model \cite{fossen2011handbook}. 
\begin{equation}
    \mathbf{M}\dot\nu + \mathbf{D}(\nu)\nu + g(\eta) = \tau_{\text{thruster}} + \tau_{\text{disturbance}}
\end{equation}
Where $\eta = [x, y, z, \phi, \theta, \psi]^T$ is the position vector, and $\nu = [u, v, w, p, q, r]^T$ is the velocity vector, each comprised of the following terms: surge, sway, heave, roll, pitch, yaw. $\mathbf{M}$ is the system inertia matrix.
\begin{equation}
    \mathbf{D}(\nu) = \mathbf{D}_l + \mathbf{D}_q(\nu)
\end{equation}
$\mathbf{D}$ denotes the hydrodynamic damping matrix, consisting of linear term $\mathbf{D}_l$ and quadratic term $\mathbf{D}_q(\nu)$. $g$ is the vector of gravitational and buoyancy forces and moments. $\tau_{\text{thruster}}$ refers to the forces and moments caused by the ASV's propulsion system, and $\tau_{\text{disturbance}}$ to the forces and moments caused by external forces such as waves and wind. The Coriolis force's impact is considered negligible within our ASV's operational speed range ($<2\ m/s$).

\section{Methodology}

\begin{figure*}[bth]
\begin{center}
\includegraphics[width=\linewidth]{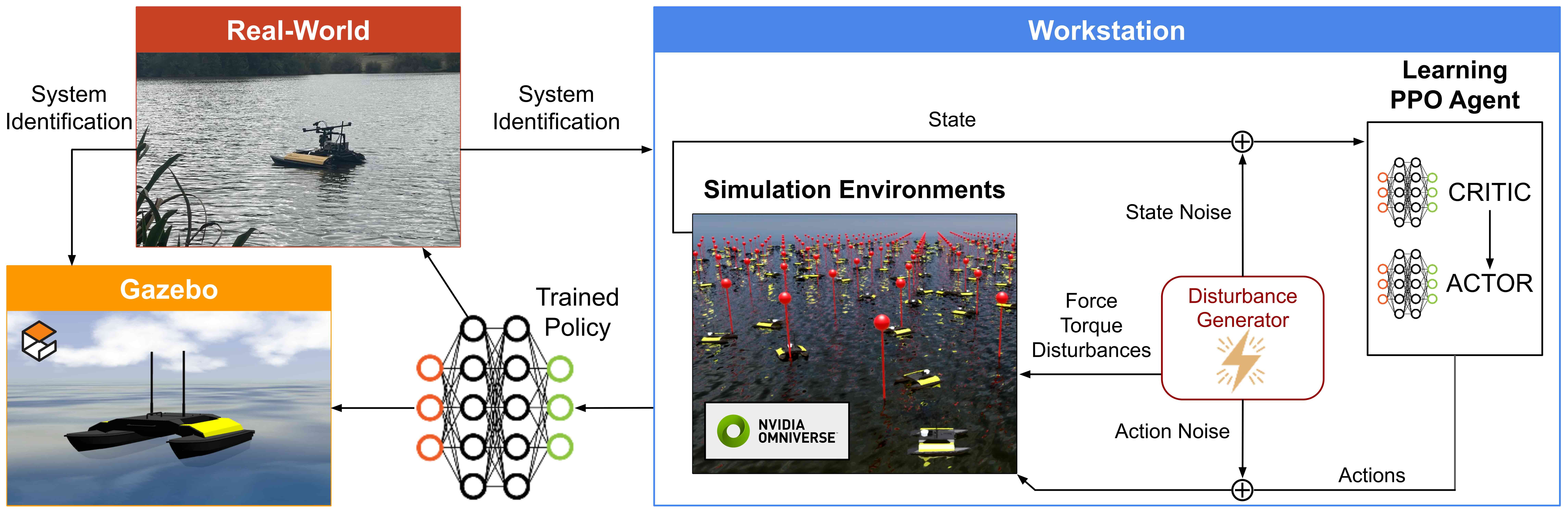}
\end{center}
   \caption{Integrated pipeline for DRL-Based ASV Control consisting of four main steps. 1. Real-world experiments for system identification; 2. Train policy in simulation; 3. Tests using an independent simulation environment; 4. Real-world tests. }
\label{fig:pipeline}
\end{figure*}

Our proposed methodology shown in Fig \ref{fig:pipeline} is composed of four main steps: System Identification, DRL training, Simulation Evaluation, and Real-world Evaluation. In the following subsection, each one is explained in more detail.

\subsection{System Identification}
\label{sub:sid}

In the system identification process, we utilized a streamlined dynamics model premised on the ASV's movement within a two-dimensional plane. This approach is substantiated by the typical operational environment of the Kingfisher ASV, which is relatively calm waters where the effects of roll and pitch on movement are minimal and thus can be disregarded for 2D motion considerations. The hydrodynamic damping matrix $\mathbf{D}$ has been tailored as follows.
\begin{equation}
\mathbf{D}_l =
\begin{bmatrix}
X_{u} & 0 & 0 \\
0 & Y_{v} & 0 \\
0 & 0 & N_{r} \\
\end{bmatrix}
\end{equation}

\begin{equation}
\mathbf{D}_q(\nu) =
\begin{bmatrix}
X_{u|u|} | u | & 0 & 0 \\
0 & Y_{v|v|} | v | & 0 \\
0 & 0 & N_{r|r|} | r | \\
\end{bmatrix}
\end{equation}
where $X_{u}$, $Y_{u}$, and $N_{r}$ are the hydrodynamic damping coefficients in surge, sway, and yaw, respectively, capturing the linear relationship. The terms $X_{u|u|}$, $Y_{v|v|}$, and $N_{r|r|}$ represent quadratic damping components that scale with the square of the velocities. For the ASV, system identification tests were conducted to derive the hydrodynamic coefficients as described in \cite{sarda2016station}. 

\begin{figure}[h!]
\begin{center}
\includegraphics[width=\linewidth]{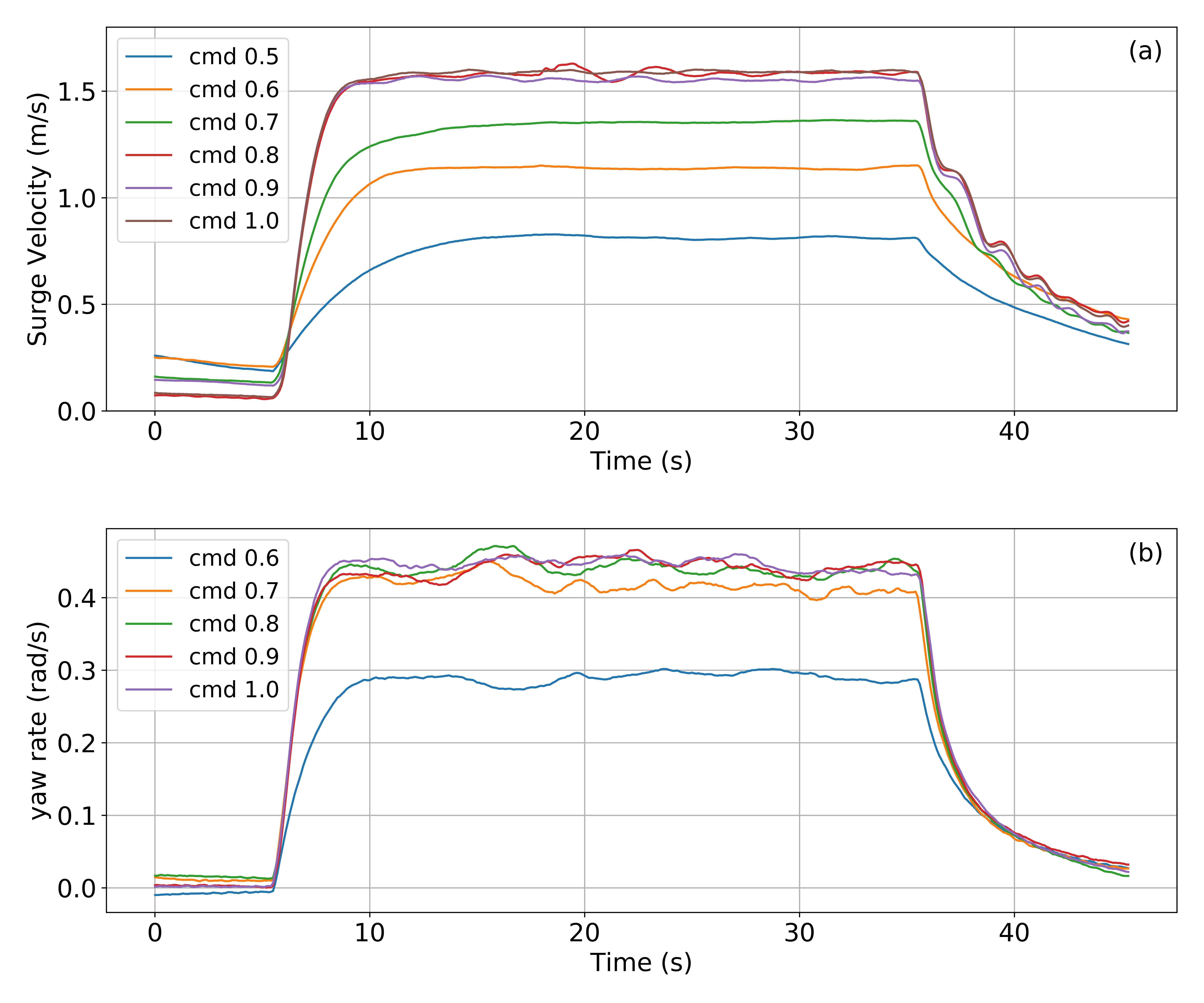}
\end{center}
\caption{System Identification Experiments. (a) shows surge velocity of the ASV during the acceleration test, and (b) shows the yaw rate of the ASV during the rotation test}
\label{fig:system_identification}
\end{figure}

Fig. \ref{fig:system_identification} (a) shows the surge direction velocity during the acceleration test with various thruster commands, which were implemented to identify surge drag coefficients. It accelerates the ASV from rest and observes its maximum velocity after reaching a steady state. In the steady state, the force generated by the thrust in the surge direction balances out with the damping force, establishing equilibrium. Fig. \ref{fig:system_identification} (b) shows the yaw rate during the rotation test, which was implemented to identify yaw drag coefficients. In this test, balanced forces were applied to both thrusters so that the ASV could rotate without surge and sway at a single point. The test results were then employed to perform quadratic curve fitting, from which we derived the drag coefficients. For the sway coefficient, circle and zigzag tests were performed. These tests consistently showed the sway velocity remaining below $0.1\ m/s$, indicating a minor influence under our test conditions. Consequently, the sway damping coefficient was manually adjusted to a high value, ensuring the model aligns with observed behaviors in the real world. The final results are shown in Tab. \ref{tab:system_parameters}.

\begin{table}[h!]
\centering
\caption{Hydrodynamics Parameters}
\renewcommand{\arraystretch}{1.1} 
\begin{tabular}{|c|c|c|}
\hline
\textbf{Parameters} & \textbf{Nominal} & \textbf{Identified} \\  
\hline
$X_{u}$     & 16.45     & 0.00       \\
$Y_{v}$     & 15.80     & 99.99      \\
$N_{r}$     & 6.00      & 0.83       \\
$X_{u|u|}$  & 2.94      & 17.26      \\
$Y_{v|v|}$  & 2.76      & 99.99      \\
$N_{r|r|}$  & 5.00      & 17.34      \\ 
\hline
\end{tabular}
\label{tab:system_parameters}
\end{table}

\begin{figure}[h!]
\begin{center}
\includegraphics[width=\linewidth]{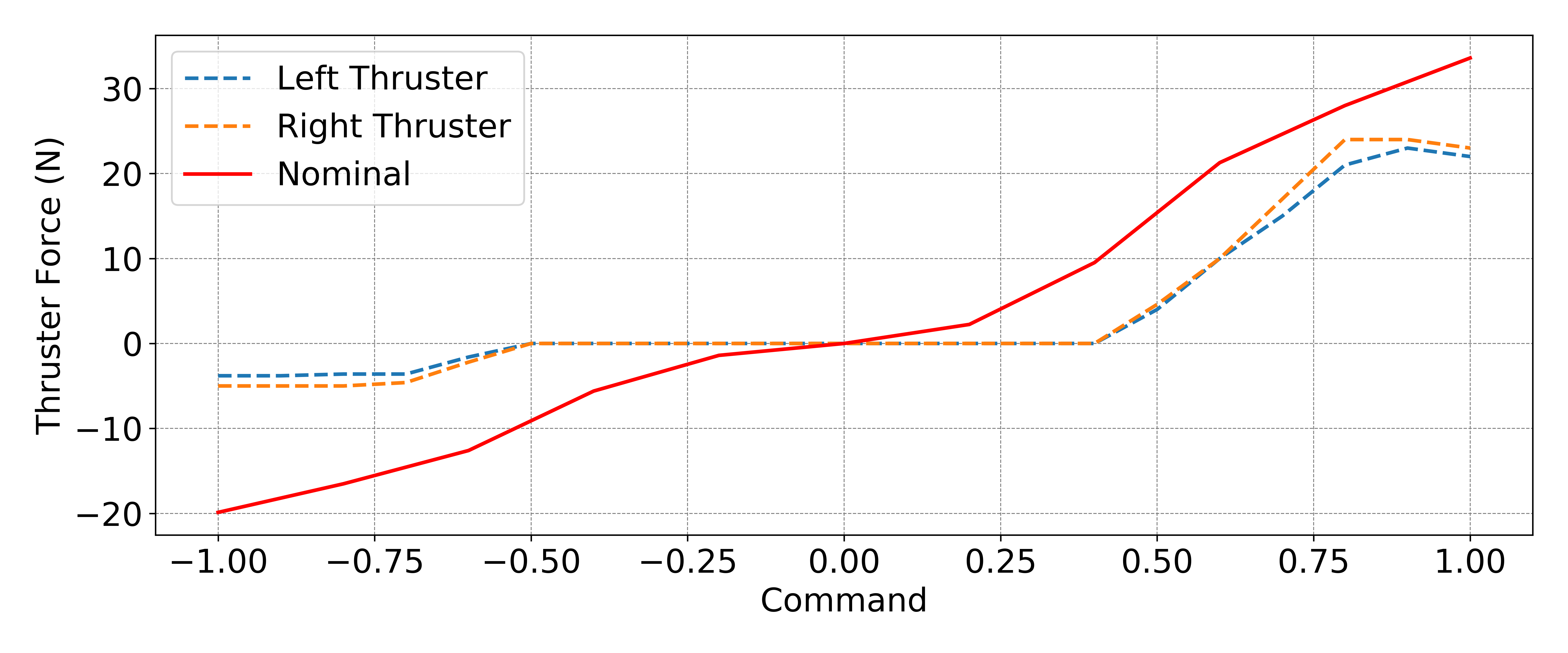}
\end{center}
\caption{Comparison of Nominal and Measured Thruster Forces}
\label{fig:thruster_forces}
\end{figure}

Fig. \ref{fig:thruster_forces} illustrates the thruster forces measured through a bollard pull test. The red line represents the nominal thruster force for the Heron ASV as provided by Clearpath Robotics. In contrast, the blue and orange lines represent the actual measured forces for the left and right thrusters, respectively, corresponding to the applied commands. It is noteworthy that the graph reveals a minor disparity between the performance of the two thrusters, which may pose challenges in learning precise control for the ASV due to its asymmetry.

\subsection{RL Framework for ASV}

Expanding upon the established methodologies from \cite{el2023rans} and \cite{richard2021heron, richard2022learning}, our work centers on developing an RL framework tailored for large-scale ASVs simulation. In this enhanced training environment, we incorporate buoyancy and hydrodynamics into IsaacSim to simulate realistic marine conditions. In particular, we performed a GPU accelerated port of the UUV gazebo plugin~\cite{manhaes2016uuv} to Isaac Sim. This port allows, for instance, the ability to compute the hydrodynamic forces of thousands of systems seamlessly on GPU. Similarly to \cite{el2023rans}, this simulator is built on top of Nvidia's Isaac Sim and OmniIsaacGym~\cite{makoviychuk2021isaac}, a simulator capable of running thousands of simulation environments in parallel. This framework marks a novel contribution to ASV development, facilitating the training and evaluation of RL agents.
Moreover, thanks to its increased training speed and broader exploration capabilities, the proposed framework offers a significant leap in training capabilities for ASV systems. It allows model-free models such as PPO~\cite{schulman2017proximal} to compete against MPC or Optimal Controllers.
Finally, by providing an open-source scalable and robust platform, this work could serve as a key step towards advanced autonomous marine operations, potentially impacting environmental sustainability efforts.

\subsection{Training process}

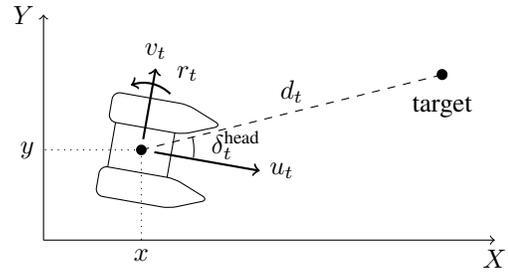
\begin{figure}[bth]
    \centering
    \usetikzlibrary{calc, angles, quotes}

\newcommand{\drawboat}[1][]{
    \begin{scope}[#1]
        \begin{scope}[shift={(-0.5,0)}] 
            \draw[rounded corners=1mm] 
            (-0.2,-0.5) -- (-0.2,0.5) -- (0,1) -- (0.2,0.5) -- (0.2,-0.5) -- cycle;
        \end{scope}
        \begin{scope}[shift={(0.5,0)}] 
            \draw[rounded corners=1mm] 
            (-0.2,-0.5) -- (-0.2,0.5) -- (0,1) -- (0.2,0.5) -- (0.2,-0.5) -- cycle;
        \end{scope}
        \begin{scope}[shift={(0,-0.4)}] 
            \draw[] (-0.3,0) -- (0.3,0);
        \end{scope}
        \begin{scope}[shift={(0,0.4)}] 
            \draw[] (-0.3,0) -- (0.3,0);
        \end{scope}
        \node at (0,0) (base_link) {$.$};
        \fill (0,0) circle (2pt); 

        \draw[->, thick] (base_link) -- ++(0,1.6) node[anchor=west] (u_arrow) {$u_{t}$};
        \draw[->, thick] (base_link) -- ++(-1.1,0) node[anchor=south] {$v_{t}$};
        
        \coordinate (v_tip) at ($(base_link)+(-0.8,0.25)$);
        \draw[->, thick] (v_tip) arc (145:210:0.5) node[pos=0.1, anchor=south west] {$r_{t}$};
    \end{scope}
}

\begin{tikzpicture}[scale=1.0] 
    \draw[->] (0,0) -- (6,0) node[anchor=north] {$X$};
    \draw[->] (0,0) -- (0,3) node[anchor=east] {$Y$};

    \drawboat[xshift=1.3cm, yshift=1.2cm, rotate=-100] 
    \coordinate (base_link) at (1.3,1.2); 
    \coordinate (T) at (5.3,2.2);
    \node at (T) {$\circ$};
    \fill (T) circle (2pt);
    \node[below of=T, node distance=12pt] {target}; 
    
    \draw[dotted] (base_link) -- (base_link |- 0,0) node [at end, below] {$x$};
    \draw[dotted] (base_link) -- (base_link -| 0,0) node [at end, left] {$y$};
    
    \draw[dashed] (base_link) -- (T);
    \path (base_link) -- (T) node [midway, above] {$d_{t}$};

    \coordinate (u_tip) at ($(base_link)+(2.0,-0.3)$);
    \pic [draw, -, "$\delta_{t}^\text{head}$", angle eccentricity=1.8, angle radius=0.7cm] {angle = u_tip--base_link--T};

\end{tikzpicture}
    \caption{Diagram of the ASV's observation space components}
    \label{fig:dynamic_model_diagram}
\end{figure}

\textbf{Observation space: }
The observation space, denoted as $\mathbf{o}_t \in \mathbb{R}^{6}$ for each time step $t$, comprises a vector of kinematic and positional features. 
\begin{equation}
    \mathbf{o}_t = [u_{t} \quad v_{t} \quad r_{t} \quad \cos(\delta_t^{\text{head}}) \quad \sin(\delta_t^{\text{head}}) \quad d_{t}]^T
\end{equation}
The components are the surge speed $u_t$, sway speed $v_t$, and yaw rate $r_t$, which represent the ASV's linear and angular velocities in the body frame. The ASV's orientation towards the goal was represented by $cos(\delta_t^{\text{head}})$ and $\delta_t^{\text{head}}$ ensuring a continuous representation of the target bearing. The last element $d_{t}$ represents the distance to the goal at time step $t$ (Fig. \ref{fig:dynamic_model_diagram}). This approach of focusing only on the local frame enhances the model's generalizability.

\textbf{Action space: }
The action space, represented as $u_t = [u_{t}^{\text{left}} \quad u_{t}^{\text{right}}]^T$, comprises the command inputs for the left and right thrusters respectively, at time step $t$. These commands are converted into thrust forces using the characteristic force curves obtained from system identification experiments. The resultant forces are then applied to the corresponding thrusters in the simulation.

\textbf{Reward function: }
To effectively train the ASV agent for the precision capture task, a compound reward function $r_t$ is used to encourage desired behaviors.
\begin{equation}
    r_t = r_t^\text{dist} + r_t^\text{head} + r_t^\text{energy} + r_t^\text{alpha} + r_t^\text{time} + r_t^\text{goal}
\end{equation}
Each reward term is detailed as follows:
\begin{equation}
    r_t^\text{dist} = \lambda_{1} \cdot (d_{t-1} - d_{t})
\end{equation}
$r_t^\text{dist}$ encourages the reduction of the distance to the goal and reflects the ASV's progress towards its goal.
\begin{equation}
    r_t^\text{head} = \lambda_{2} \cdot (e^{k_{1} (\delta_t^\text{head})^4} + e^{k_{2} (\delta_t^\text{head})^2})
\end{equation}
$r_t^\text{head}$ promotes alignment of the ASV's heading with the goal orientation with continuous reward with a pronounced peak near zero.
\begin{equation}
    r_t^\text{energy} =\lambda_{3} \cdot (e^{k_{3} E} - 1)
\end{equation}
$r_t^\text{energy}$ term penalizes excessive use of actuation to encourage energy efficiency, where $E = (u_{t}^{\text{left}})^2 + (u_{t}^{\text{right}})^2$.
\begin{equation}
    r_t^\text{alpha} = \lambda_{4} \cdot (e^{k_{4} |r_{t-1} - r_{t}|} - 1)
\end{equation}
$r_t^\text{alpha}$ discourages abrupt changes in angular velocity, thereby facilitating stable and gradual turns.
\begin{equation}
    r_t^\text{time} = \lambda_{5}
\end{equation}
$r_t^\text{time}$ applies a penalty for each time step, encouraging the agent to reach the goal quickly. 
\begin{equation}
    r_t^\text{goal} =
    \begin{cases} 
        \lambda_{6}, & \text{if } d_{t} < d_{threshold} \\
        0, & \text{otherwise}
    \end{cases}
\end{equation}
Finally, the $r_t^\text{goal}$ component grants a one-time reward when the ASV successfully reaches the goal. $d_{\text{threshold}}$ determines the required proximity for successful goal achievement. 
$\lambda_1, \lambda_2, \lambda_3, \lambda_4, \lambda_5, \lambda_6$ are weights for each reward component, and $k_1, k_2, k_3$ and $k_4$ are constants that adjust the sensitivity of the reward components to their respective errors.

\textbf{Training Details: }

\begin{table}[ht]
\centering
\caption{The reward function's parameter values}
\renewcommand{\arraystretch}{1.3} 
\begin{tabular}{c c c c c c c}
\hline
Weights & $\lambda_1$ & $\lambda_2$ & $\lambda_3$ & $\lambda_4$ & $\lambda_5$ & $\lambda_6$\\ 
Values & 1.0 & 0.02 & 0.01 & 1.0 & -0.2 & 30.0\\
\hline
Parameters & $k_1$ & $k_2$ & $k_3$ & $k_4$ & $d_{\text{threshold}}$ \\ 
Values & -10.0 & -0.1 & 1.0 & -0.33 & 0.1 m \\
\hline
\end{tabular}
\label{tab:reward_function_parameters}
\end{table}

The values and parameters of the reward function, shown in Tab. \ref{tab:reward_function_parameters}, have been carefully calibrated to optimize the agent behavior. The cumulative reward per time step remains negative prior to the goal being reached, thereby refining the ASV's approach toward the target. A conservative lower value for $d_{\text{threshold}}$ was chosen, which is less than the actual required proximity to capture the target. 

Tab. \ref{tab:rl_config} presents the core learning configurations employed in our training. The training process was completed with an Nvidia GeForce GTX 3090 GPU with 24GB RAM for approximately 10 minutes.

\begin{table}[!ht]\centering
\caption{Reinforcement learning configurations}\label{tab: }
\renewcommand{\arraystretch}{1.1} 
\scriptsize
\begin{tabular}{c c}\toprule
Parameters & Values \\\midrule
actor neural network & 128 $\times$ 128 \\
critic neural network & 128 $\times$ 128 \\
learning rate & 0.0001 \\
gamma & 0.99 \\
number of environments & 1024 \\
sample batch size & 16384 \\
max iterations & 1000 \\
max steps per episode & 3000 \\
time step size & 0.02 \\
\bottomrule
\end{tabular}
\label{tab:rl_config}
\end{table}

To robustly train the DRL agent, we incorporated a comprehensive randomization strategy across the simulation environment. This strategy involved introducing observation noise to replicate the accuracy of sensors employed in the ASV (position: $\pm3\ cm$, orientation: $\pm0.025\ rad$) and perturbing action commands to reflect control imprecision. $\pm10\%$ random variation in the drag coefficients was introduced to account for uncertainties in its hydrodynamic model, thereby diversifying the training environments and preventing overfitting to idealized hydrodynamic parameters.

We modeled external force and torque disturbances to simulate natural environmental conditions, applying a force field with a random offset of $\pm2.5\ N$ and a maximum sine force amplitude of 2.5 N, along with torque disturbances having a random offset of $\pm1.0\ N\cdot m$ and a sine torque amplitude of $1.0\ N\cdot m$. These disturbances, representing about 10\% of the ASV's thrust capacity, are significant enough to test control strategy robustness without overwhelming the system, enhancing agents' reliability and performance in actual operational scenarios. 

Additionally, we applied up to 50\% randomization to thruster forces to account for fluctuations due to battery level variations and discrepancies between the nominal and actual measured thruster forces, ensuring the training process encompasses a realistic spectrum of operational conditions.

\subsection{Simulation Environment}
The initial evaluation of the trained models was conducted in the Gazebo simulation environment, using the UUV \cite{manhaes2016uuv} plugin. This choice was motivated by the chance to validate the models' performance independence from Isaac Sim training environment and to facilitate ROS integration for field tests. Gazebo's parameters, calibrated with system identification parameters identified in Sec. \ref{sub:sid}, provided an environment for assessing model behavior.

\subsection{Autonomous Surface Vehicle Platform Integration}

For the field test experiments, the Kingfisher ASV, developed by Clearpath Robotics, was used. It features a catamaran design, with dimensions of $1.35\ m$ in length and $0.98\ m$ in width with weight of $35.96\ Kg$. Its propulsion system is based on a thruster mounted on each hull, as shown in Fig. \ref{fig:asv-real}. 
The motors are driven by a proprietary low-level controller board that controls each thruster using an input command of at least $10\ Hz$ between $[-1.0, 1.0]$.
As a higher-level processing unit, we used an NVIDIA Jetson Xavier with CUDA integrated into the ASV to provide energy-efficient high-performance computing capable of providing real-time inference for the RL agent. 
Power was provided using a 4-cell LiPo battery pack with $22\ Ah$ capacity.
Despite being equipped with two cameras and a laser, these sensors were not used. For experimental purposes, perception was abstracted, and the goal was provided programmatically. The localization relied on a high precision SBG Ellipse-D integrated IMU and RTK GPS with a dual-antenna, multi-ban GNSS receiver providing high position, velocity, and heading precision of $0.02\ m$, $0.03\ m/s$, and $0.5^{\circ}$ respectively.
\section{Experiments}

To evaluate how the integration of system identification with domain randomization affects the model's performance, experiments were conducted in simulation and real-world environments with agents trained using the nominal values (NV) or identified values (SID) as described in Tab. \ref{tab:system_parameters}. Thruster force domain randomization (DR) was applied in some cases, resulting in four different tests (Tab. \ref{tab:tested-agents}).

\begin{table}[!htp]
\centering
\caption{Agents training configurations for experiments}
\label{tab:tested-agents}
\scriptsize
\begin{tabular}{l|c|cc}\toprule
   &Thruster and  &Thruster Force  \\
Name & Hydrodynamics parameters             & Randomization \\\midrule
NV &Nominal &0\% \\
NV-DR &Nominal &50\% \\
SID &Identified &0\% \\
SID-DR &Identified &50\% \\
\bottomrule
\end{tabular}
\end{table}

\subsection{Simulation Evaluation}

\begin{figure}[bth]
\begin{center}
\includegraphics[width=\linewidth]{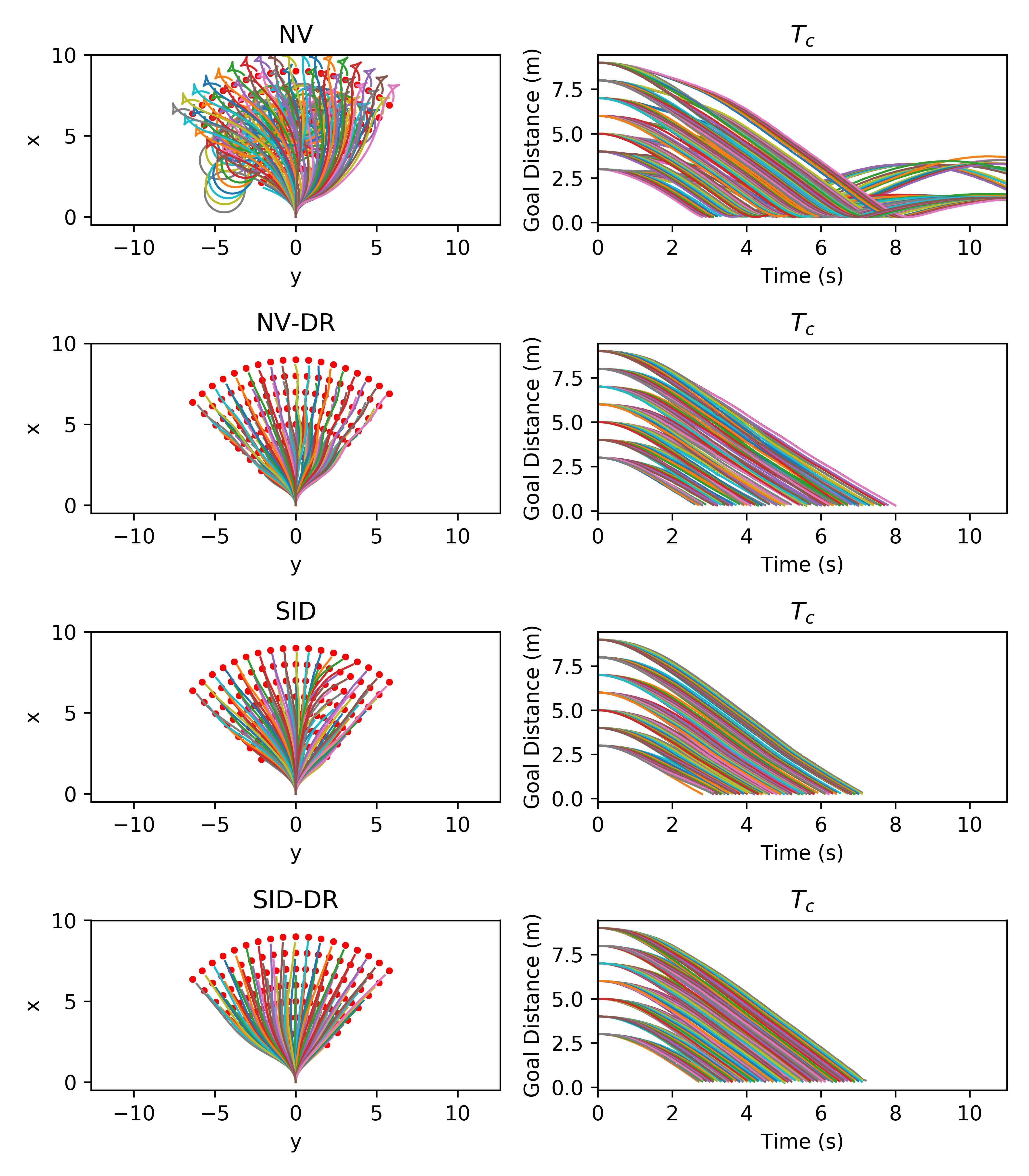}
\end{center}
\vspace{-5pt}
   \caption{Simulation tests showing trajectories (left) and goal distance over time (right). NV exhibits difficulty in reaching the target (red dot), while SID-DR completes the task faster.}
\vspace{-10pt}

\label{fig:sim_eval}
\end{figure}

Target locations were generated with distances set between 3 to 9 meters at 1-meter intervals and angles from -45 to 45 degrees in 5-degree steps. This setup enabled a focused evaluation of navigational efficiency necessary for capturing waste. The Simulation results, detailed in Figure \ref{fig:sim_eval} show that NV struggled to reach the target in the first approach, having to turn around and correct the trajectory, while other agents presented better results.

\subsection{Real-world Evaluation}

\begin{figure}[bth]
\begin{center}
\includegraphics[width=\linewidth]{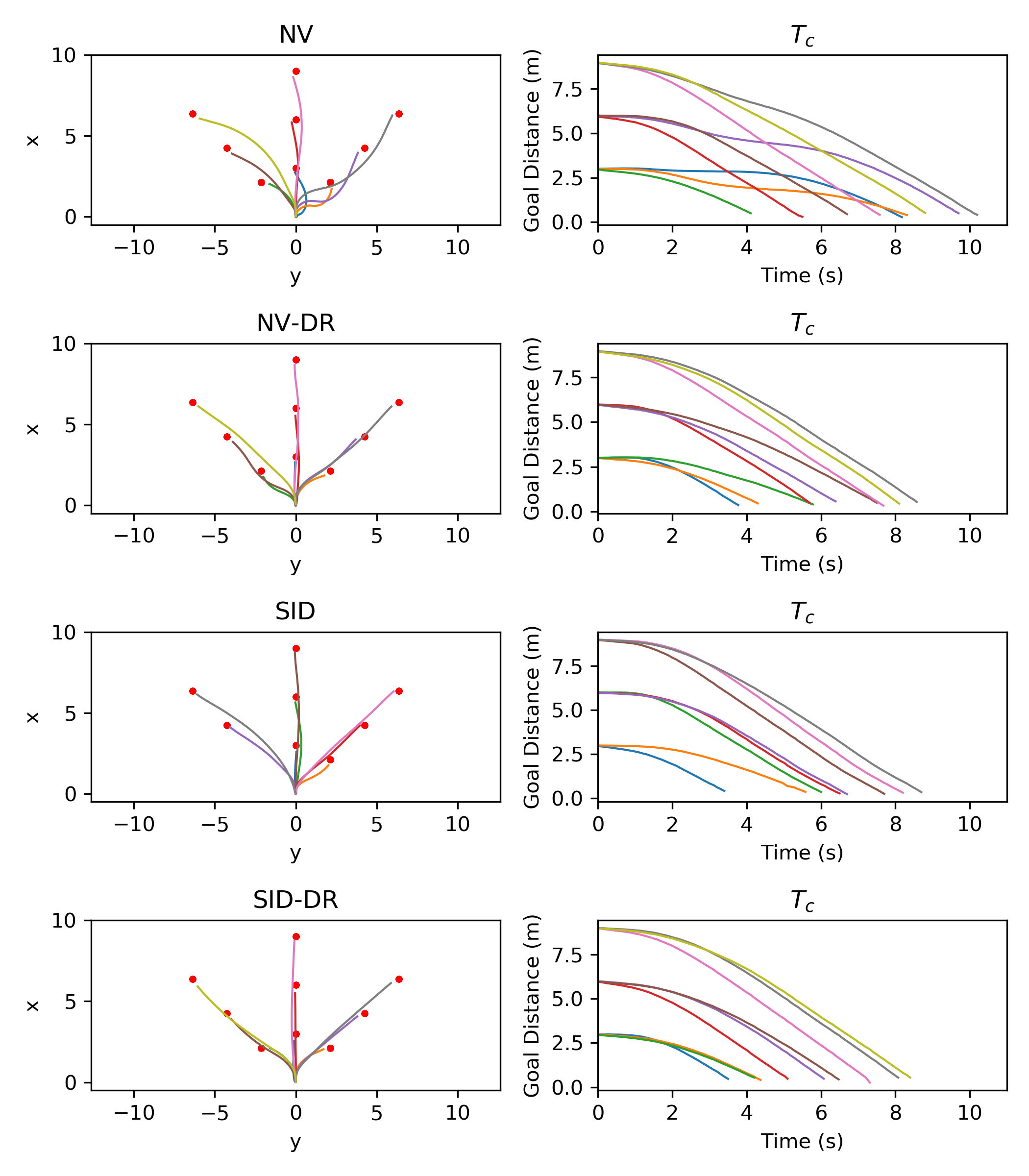}
\end{center}
\vspace{-5pt}
   \caption{Field tests demonstrate that all models are capable of reaching the target while SID-DR completes the task faster with smoother trajectories.}
\vspace{-10pt}
\label{fig:field_eval}
\end{figure}

To validate simulations and assess model applicability, field tests were conducted on a day with mild wind speeds (reported maximum of 6 km/h) per local weather. Focusing on controlled yet realistic conditions, these tests covered a subset of predefined simulation goals due to battery constraints. Results are shown in Fig. \ref{fig:field_eval} where SID-DR provides smoother trajectories while reaching the target faster.

\subsection{Evaluation Metrics}

Each model's performance was evaluated based on two main metrics: completion time ($T_c$) and accumulated energy consumption $E_{acc}$. In general, thruster power consumption is positively correlated to its thrust. Hence, the accumulated power consumption $E_{acc}$ of the controller can be represented by Eq. \ref{eq:accenergy}, where $N$ is the number of the data points and $a_l^{i}$ and $a_r^{i}$ are the action commands on thruster left and right.

\begin{eqnarray}\label{eq:accenergy}
E_{\text{acc}} \propto {\sum_{i=1}^{N}(|a_l^{i}|+|a_r^{i}|)} 
\end{eqnarray}

\begin{figure}[bth]
\begin{center}
\includegraphics[width=\linewidth]{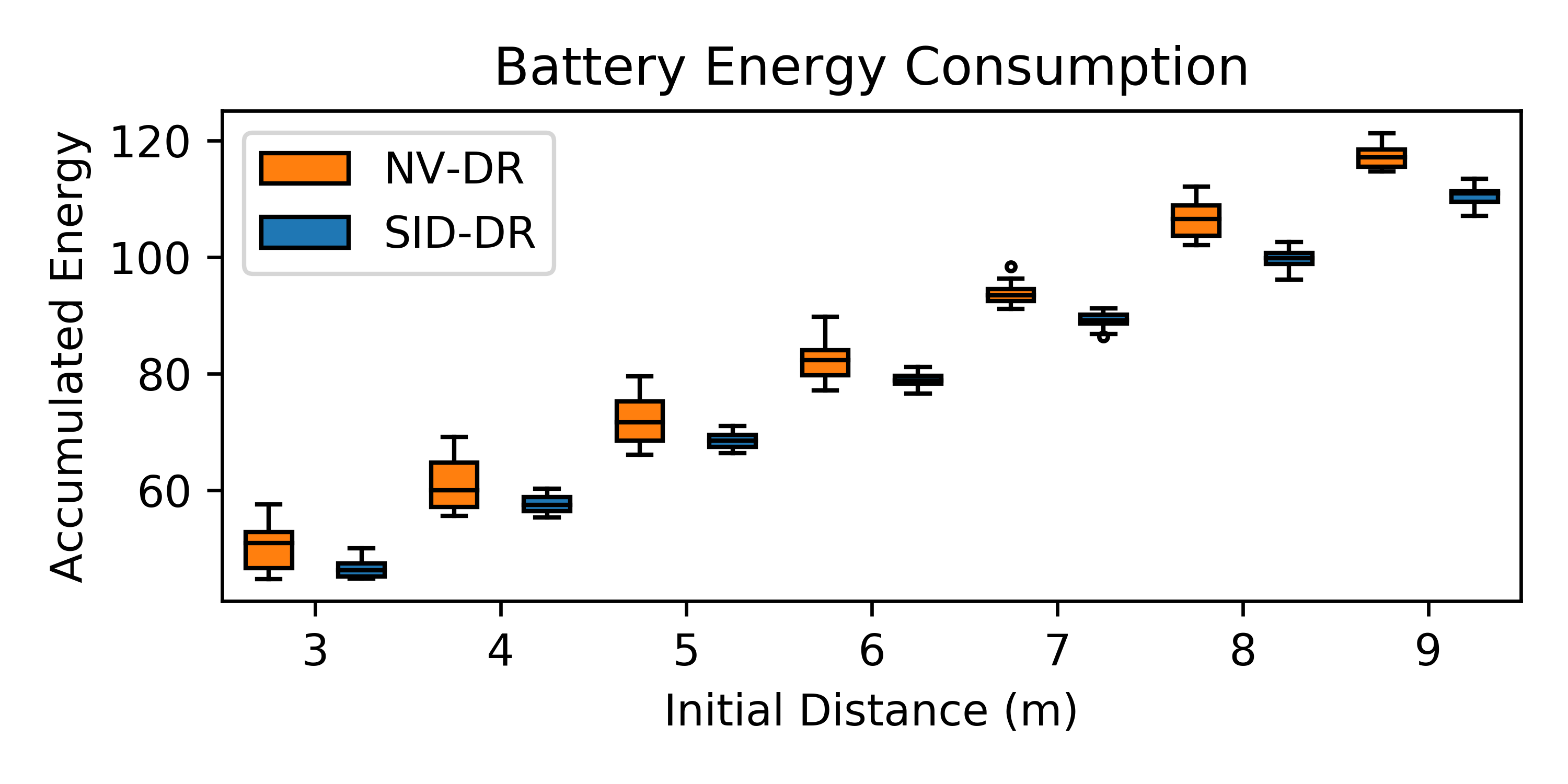}
\end{center}
\vspace{-15pt}
   \caption{Large-scale Simulation Analysis Shows that SID-DR Outperforms NV-DR in Battery Efficiency.}
\label{fig:energy_eval}
\vspace{-15pt}
\end{figure}

\subsection{Results}

\begin{table}[!htp]\centering
\caption{Time and Energy metrics}
\label{tab:metrics_comparison}
\scriptsize
\begin{tabular}{lrrrr|rrrr}\toprule
& &\multicolumn{3}{c}{$T_c (s)$} &\multicolumn{3}{c}{$E_{acc}$} \\\cmidrule{3-8}
& Goal distance &$3m$ &$6m$ &$9m$ &$3m$ &$6m$ &$9m$ \\\midrule
\multirow{4}{*}{\rotatebox[origin=c]{90}{\textbf{Sim}}} &NV &4.0 &15.4 &16.9 &63.2 &236.6 &269.8 \\
&NV-DR &3.2 &5.3 &7.3 &50.2 &82.3 &117.5 \\
&SID &3.3 &5.0 &6.9 &44.7 &81.0 &116.1 \\
&SID-DR &3.0 &4.9 &6.7 &46.6 &78.9 &110.5 \\
\midrule
\multirow{4}{*}{\rotatebox[origin=c]{90}{\textbf{Real}}} &NV &6.9 &7.3 &8.9 &105.8 &114.9 &139.1 \\
&NV-DR &4.6 &6.5 &8.1 &77.8 &105.4 &131.9 \\
&SID &4.5 &6.4 &8.2 &74.9 &88.6 &123.3 \\
&SID-DR &4.0 &5.9 &7.9 &57.8 &90.8 &125.2 \\
\bottomrule
\end{tabular}
\end{table}

Our field experiments demonstrated a $13.1\%$ reduction in energy consumption and a $7.4\%$ improvement in task completion efficiency when comparing the performances of NV-DR and SID-DR approaches. These numbers are extracted from the average across the different initial distances presented on Tab. \ref{tab:metrics_comparison} and underscore the efficacy of incorporating SID in conjunction with DR to optimize autonomous operations.

Real-world conditions, characterized by dynamic elements such as waves and wind, naturally require more energy and time to complete the task than the simulated environments test, where such disturbances were absent. This can also be observed on Tab. \ref{tab:metrics_comparison}. 

The data points gathered from field tests are comparatively limited. Therefore, the evaluation of energy consumption was also conducted in simulation to obtain metrics of greater statistical significance. The results shown in Fig. \ref{fig:energy_eval} corroborate the trends observed in real-world tests, specifically when comparing NV-DR and SID-DR across various initial distances.

\section{Discussion}

The field tests of DRL agents for autonomous waste collection vessels revealed critical challenges rarely observed in simulation. The initial issue was the significant discrepancy between simulated and real-world performance during preliminary trials, which highlighted the need for improved system identification and domain randomization techniques presented in this work. Primarily, adjusting for thruster dynamics was crucial to bridge the reality gap and enhance agent adaptability. In addition, the nonlinear relationship between battery charge levels and motor power output significantly affected preliminary results. External disturbances and sensor noise further compounded these challenges, necessitating a minimal domain randomization level to train agents capable of generalizing to real-world deployment. Finally, the intricacies of reward shaping proved to be a substantial hurdle. Developing a reward function that effectively guides the agent towards efficient waste collection while avoiding local optima requires careful calibration and thoughtful consideration.

\section{Conclusion}

In this study, we contribute to the field of marine robotics by enhancing a highly parallelizable RL framework with the integration of buoyancy and hydrodynamic models. Furthermore, we share a methodology that combines system identification with domain randomization, effectively narrowing the sim-to-real gap in ASV simulations. Our approach is validated through both simulation and real-world trials, specifically focusing on the task of capturing floating waste. Results demonstrate not only a reduction in task completion time but also a decrease in energy consumption. We are optimistic that the dissemination of our findings and open implementation can empower the research community, contributing to the development of more efficient and versatile autonomous systems in aquatic environments.



\bibliographystyle{IEEEtran}
\bibliography{IEEEabrv,references}

\end{document}